\documentclass[runningheads]{llncs}
\usepackage[T1]{fontenc}
\usepackage{amssymb}
\usepackage{appendix}
\usepackage{graphicx}
\usepackage{multirow}
\usepackage{subcaption}

\begin{document}
\title{WineGraph: A Graph Representation For Food-Wine Pairing}
%
%\titlerunning{Abbreviated paper title}
% If the paper title is too long for the running head, you can set
% an abbreviated paper title here
%Zuzanna Gawrysiak, Agata Żywot, Agnieszka Ławrynowicz

\author{Zuzanna Gawrysiak\inst{1}\orcidID{0009-0005-7674-5783} \and
Agata Żywot\inst{1}\orcidID{0009-0002-2851-4866} \and
Agnieszka Ławrynowicz\inst{1}\orcidID{0000-0002-2442-345X}}
\authorrunning{Z. Gawrysiak et al.}
% First names are abbreviated in the running head.
% If there are more than two authors, 'et al.' is used.
%
\institute{  Poznan University of Technology\\
  Faculty of Computing and Telecommunications\\
  Piotrowo 3, 60-965 Poznań, Poland\\
\email{\{zuzanna.gawrysiak,agata.zywot\}@student.put.poznan.pl\\
%  agata.zywot@student.put.poznan.pl\\
  alawrynowicz@cs.put.poznan.pl}\\
%\url{http://www.springer.com/gp/computer-science/lncs} \and
%ABC Institute, Rupert-Karls-University Heidelberg, Heidelberg, Germany\\
%\email{\{abc,lncs\}@uni-heidelberg.de}
}
\maketitle              % typeset the header of the contribution
\begin{abstract}
We present \textit{WineGraph}, an extended version of \textit{FlavorGraph}, a heterogeneous graph incorporating wine data into its structure. This integration enables food-wine pairing based on taste and sommelier-defined rules. Leveraging a food dataset comprising 500,000 reviews and a wine reviews dataset with over 130,000 entries, we computed taste descriptors for both food and wine. This information was then utilised to pair food items with wine and augment FlavorGraph with additional data. The results demonstrate the potential of heterogeneous graphs to acquire supplementary information, proving beneficial for wine pairing.

\end{abstract}
\section{Introduction}
The field of food and wine pairing has garnered significant attention in recent years, with various studies focusing on understanding the intricate relationships between flavours and aromas~\cite{wine-pairing-study}. Despite the wealth of information available on food and wine individually, there is a noticeable gap in comprehensive datasets specifically dedicated to food-wine pairing~\cite{DBLP:conf/nips/BenderSKHHHBW23}. %This gap poses a challenge for researchers and enthusiasts alike, as the lack of structured data hinders the development of effective algorithms and models for wine and food recommendations.
In light of recent advancements in food recommendation and substitution using graphs, the primary objective of this research is to enhance the existing heterogeneous graph structure, FlavorGraph~\cite{flavor-graph}, by incorporating detailed information about wine.

FlavorGraph is a large-scale graph network comprising food and chemical compound nodes.
Recent advancements in graph-based embedding approaches, particularly utilizing node2vec~\cite{grover2016node2vec} and metapath2vec~\cite{DBLP:conf/kdd/DongCS17} algorithms, have shown promise in constructing conceptual representations from network data. These methods generate random walks analogous to sentences in word2vec~\cite{mikolov2013distributed}, leveraging relations within the network. The application of these techniques to food pairing recommendations involves constructing representations of foods based on relationships between ingredients and chemical compounds.

In this work, we present \emph{WineGraph}, an extended version of FlavorGraph, a heterogeneous graph incorporating wine data into its structure.

\section{Preliminaries} 
%\subsection{FlavorGraph}

%\subsection{Metapath2vec Framework}
Building upon graph-based embedding approaches, and resources: Recipe1M~\cite{marin_recipe1m_2019}, FlavorDB~\cite{garg_flavordb_2018}, and HyperFoods~\cite{veselkov_hyperfoods_2019}; FlavorGraph uses metapath2vec to generate conceptual representations of food. By defining food-specific metapaths and considering both chemical and statistical aspects of food pairing, the model addresses optimization challenges arising from the sparse availability of chemical information for food ingredients.
Multiple food-specific metapaths were designed in this respect by~Park at al.~\cite{flavor-graph} to facilitate the transfer of scarce chemical information from the compound nodes to non-hub ingredient nodes via chemical-hub ingredients. These metapaths enable training on complex relations, including food-food and food-chemical compounds interactions. %The resulting food representations were utilized in various downstream tasks, including food pairing.

More formally, following~Dong et al.~\cite{DBLP:conf/kdd/DongCS17} we define \emph{heterogeneous network}~(Definition~\ref{def:heterrogeneuosnetwork}).
\vspace{-5pt}
\begin{definition}
A \emph{heterogeneous network} is defined as a graph $G = (V, E, T)$ in which each node $v$ and each link $e$ are associated with their mapping functions $\phi(v) : V \rightarrow T{_V}$ and $\phi(e) : E \rightarrow T_E$, respectively. $T_V$ and $T_E$ denote the sets of object and relation types, where $|T_V| + |T_E| > 2$.
\label{def:heterrogeneuosnetwork}
\end{definition}

Subsequently, by taking into account a heterogeneous network as our input, we formulate the task of heterogeneous network representation learning in the following manner.
\vspace{-2pt}
\begin{definition}
\emph{Heterogeneous network representation learning}: Given a heterogeneous network $G$, the task is to learn the $d$-dimensional latent representations $\mathbf{X} \in \mathbb{R}^{|V|\times d}, d \ll |V|$ that are able to capture the structural and semantic relations among them.    
\end{definition}

Park at al.~\cite{flavor-graph} extended metapath2vec model with an additional chemical structure learning layer.  
\vspace{-3pt}
\section{Materials and Methods}
\begin{figure}
    \centering   \includegraphics[scale=0.14]{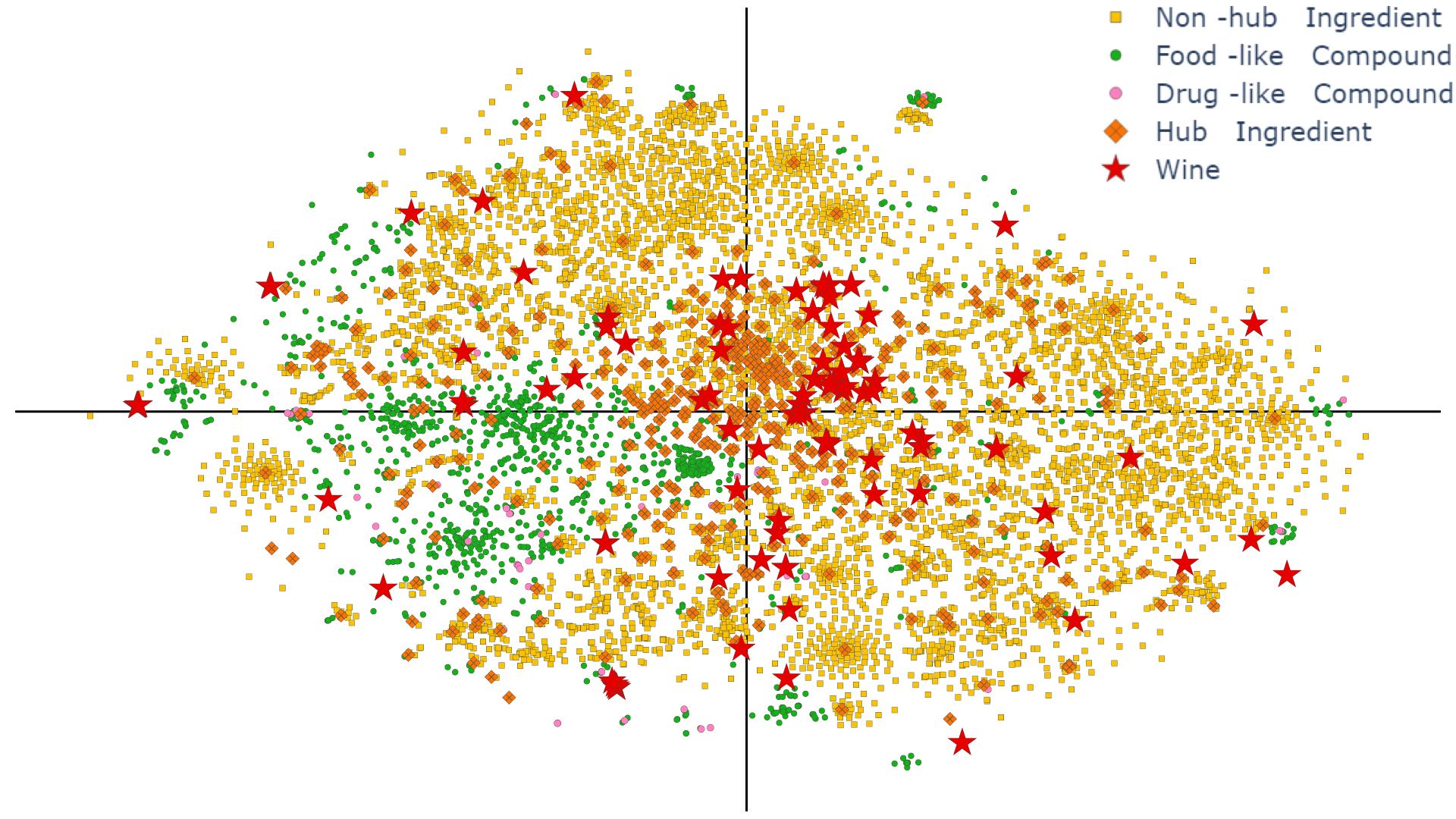}
    \caption{Visualization of the WineGraph using t-SNE projection.}
    \label{fig:winegraph}
\end{figure}

To integrate wine pairing into FlavorGraph, we have performed four key steps: 1) pre-processing food and wine review datasets, 2) calculating aroma descriptors based on resulting phrases, 3) creating a list of food-wine pairings, 4) incorporating the resulting data into FlavorGraph. 
The resulting WineGraph is visualized in Figure~\ref{fig:winegraph}).

%    \item mapping resulting phrases to predefined %aroma descriptors,
%    \item creating a list of food-wine pairings, 
%   \item incorporating the resulting data into 
%\begin{enumerate}
 
%\item pre-processing food and wine review datasets,
%    \item mapping resulting phrases to predefined %aroma descriptors,
%    \item creating a list of food-wine pairings, 
%   \item incorporating the resulting data into FlavorGraph.
%\end{enumerate}

\vspace{-10pt}
\subsection{Data Preparation}

 The first step 
%the initial step 
involved 
%pregenerating pairings. This was accomplished by 
utilising two datasets: Amazon Fine Food Reviews\footnote{\url{ https://www.kaggle.com/datasets/snap/amazon-fine-food-reviews}} and Wine Reviews\footnote{\url{https://www.kaggle.com/datasets/roaldschuring/wine-reviews}}.  The text was first tokenized and normalized. 
Then the most frequent phrases which consisted of 1-3 tokens used together most frequently (i.e., $n$-grams) were extracted, obtaining flavour descriptors like \emph{fruit flavour}, \emph{acid}, \emph{black cherries} etc.
The second step involved mapping the phrases to aroma descriptors (wine only) using the UC Davis wine wheel\footnote{\url{http://winearomawheel.com/}}. This wheel is comprised of three tiers of aromas, ranging from specific to broad, facilitating the generalization of descriptors. An illustration of these levels would be raspberry -> berry -> fruit.
In the third step, 
all preprocessed reviews were utilised to train the word2vec model. We also calculated TF-IDF embeddings.
The result was 300-dimensional aroma vectors 7 non-aroma scalers for wines and 300-dimensional aroma vectors for food.
Table~\ref{tab:core-tastes} illustrates core tastes and obtained relevant embeddings. 

\begin{table}[t]
\centering
\caption{Closest and furthest items for core tastes - cells contain food item (cosine similarity).}
\small
\begin{tabular}{@{}l|ll@{}}
\textbf{Taste} & \textbf{Closest Item} & \textbf{Furthest Item} \\ 
\hline
Weight & Pizza (0.546) & Dragonfruit (-0.118) \\
Sweet & Pineapple (0.536) & Mackerel (-0.254) \\
Acid & Tart (0.600) & Biscuit (-0.121) \\
Salt & Bacon (0.672) & Nectar (-0.122) \\
Piquant & Chili (0.434) & Sole (-0.106) \\
Fat & Cake (0.618) & Coffee (-0.115) \\
Bitter & Tart (0.475) & Platter (-0.155) \\
\end{tabular}

\label{tab:core-tastes}
\end{table}

%\vspace{-10pt}
\subsection{Pairing Procedure}
The resulting embeddings are used to generate food-wine pairings based on predefined rules, encompassing attributes like \emph{sweetness} and \emph{acidity}, defined as numerical thresholds. In the rules, we use 7 types of numerical variables that correspond to the set \{sweet, bitter, salty, acid, fatty, piquant, weight\} whose values are normalized to be in the range from 0 to 1. 
The pairing procedure is: 1) calculate aroma and non-aroma descriptors with the use of the trained word2vec model, 2) eliminate wines that do not match the food item (predefined set of rules), 3) find congruent and contrasting wines (predefined set of rules), 4) sort by aroma similarity. These steps were first proposed by Roald Schuring\footnote{\url{https://towardsdatascience.com/robosomm-chapter-5-food-and-wine-pairing-7a4a4bb08e9e}}.

%\vspace{-5pt}
A wine and food pairing must first meet all elimination/constraint rules (that is, not be rejected by any) and then any pairing rule to conclude as  ''pairing'' is true.
In Tables~\ref{tab:elimination} and ~\ref{tab:constraint} we provide two sets of such sommelier-defined rules~\cite{wine-pairing-study}\footnote{\url{https://academy.getbackbar.com/the-basics-wine-and-food-pairing}}\footnote{\url{https://cdn.courtofmastersommeliers.org/uploads/2022/11/Food-and-Wine-1.pdf}} (elimination rules and congruent rules).
We can formalize the former as constraints and the latter as decision rules~\cite{dblp:series/cogtech/FurnkranzGL12,molnar2022}.
%\vspace{-5pt} 
\begin{table}[t]
\caption{Elimination rules (constraints).}
\label{tab:elimination}
    \centering
    \small
    \begin{tabular}{p{2.5cm} p{9cm}}
        \hline
      \textbf{weight} & {\ }Wine should have at least the same ''body'' as the food. \\
      \textbf{acidity} & {\ }The wine should be at least as acidic as the food. \\
      \textbf{sweetness} & {\ }The wine should be at least as sweet as the food. \\
      \textbf{bitterness} & {\ }Bitter wines do not pair well with bitter foods. \\
      \textbf{bitter-salt} &  {\ }Bitterness and saltiness do not pair well together. \\
     \textbf{acid-bitter} & {\ }Acidity and bitterness do not pair well together. \\
    \textbf{acid-piquant} & {\ }Acidity and piquancy (spiciness) do not pair well together.
             \\
            \hline

    \end{tabular}
    \label{tab:my_label}
\end{table}
%\vspace{-5pt}
\begin{table}[t]
\caption{Congruent/contrasting rules (decision rules).}
\label{tab:constraint}
    \centering
    \small
    \begin{tabular}{p{2.5cm} p{9cm}}
            \hline
      \textbf{sweet pairing} & {\ }Sweet food is paired well with highly bitter, fatty, piquant, salty, or acidic wines. \\
      \textbf{acid pairing} & {\ }Acidic food is paired well with highly sweet, fatty, or salty wines. \\
      \textbf{salt pairing} & {\ }Salty foods are paired well with highly bitter, sweet, fatty, piquant, or acidic wines. \\
      \textbf{piquant pairing} & {\ }Spicy food is paired well with highly sweet or fatty wines. \\
      \textbf{fat pairing} &  {\ }Fatty food is paired well with highly bitter, sweet, piquant, or acidic wines. \\
     \textbf{bitter pairing} & {\ }Bitter food is paired well with highly sweet or fatty wines. \\
             \hline
    \end{tabular}
    \label{tab:my_label}
\end{table}
The sample elimination rule for acidity (the wine should be at least as acidic as the food) is shown below:
$$wine_{acid} >= food_{acid} \Rightarrow \mathbf{eliminate_{false}}$$
%\vspace{-3pt}
%acid-bitter - acidity and bitterness do not pair well together.
%\[(food_{acid} > 0.75 \land wine_{bitter} < 0.5) \lor
%(food_{bitter} > 0.75 \land wine_{acid} < 0.5)\]
and sample pairing rule 
%for sweetness pairing %(sweet food is paired well with highly bitter, fatty, piquant, salty, or acidic wines):
%\[food_{sweet} > 0.75 \land (wine_{bitter} > 0.75 \lor
%wine_{fat} > 0.75 \lor wine_{piquant} > 0.75 \lor
%wine_{acid} > 0.75)\]
for acidity (acidic food is paired well with highly sweet, fatty, or salty wines) is as follows:
$$food_{acid} > 0.75 \land (wine_{sweet} > 0.75 \lor
wine_{fat} > 0.75 \lor wine_{salt} > 0.75)\Rightarrow \mathbf{pairing_{true}}$$
%Prior to the pairing procedure, the reviews underwent tokenization, and the most frequent n-grams were extracted. 
%In the case of wine reviews, an additional step involved mapping them to aroma descriptors using the UC Davis wine wheel. 
%Subsequently, 
\subsection{Incorporating Pairing Into Knowledge Graph}
\begin{figure}[h]
\centering
\begin{subfigure}{.42\textwidth}
  \centering
  \includegraphics[width=0.99\linewidth]{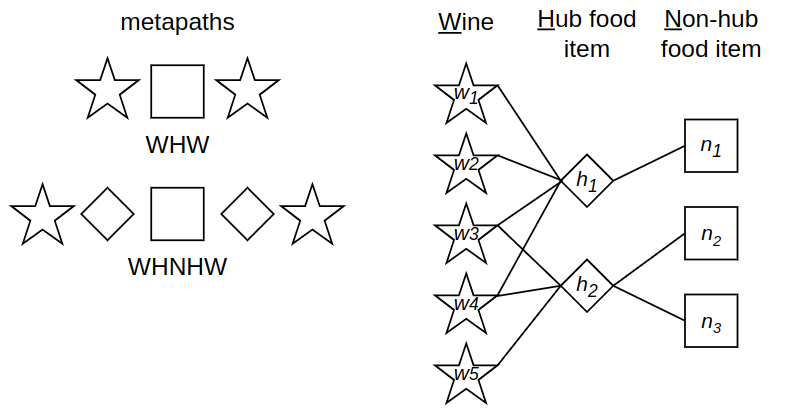}
  \caption{A food and wine network.}
  \label{fig:network}
\end{subfigure}%
\begin{subfigure}{.55\textwidth}
  \centering
  \includegraphics[width=0.99\linewidth]{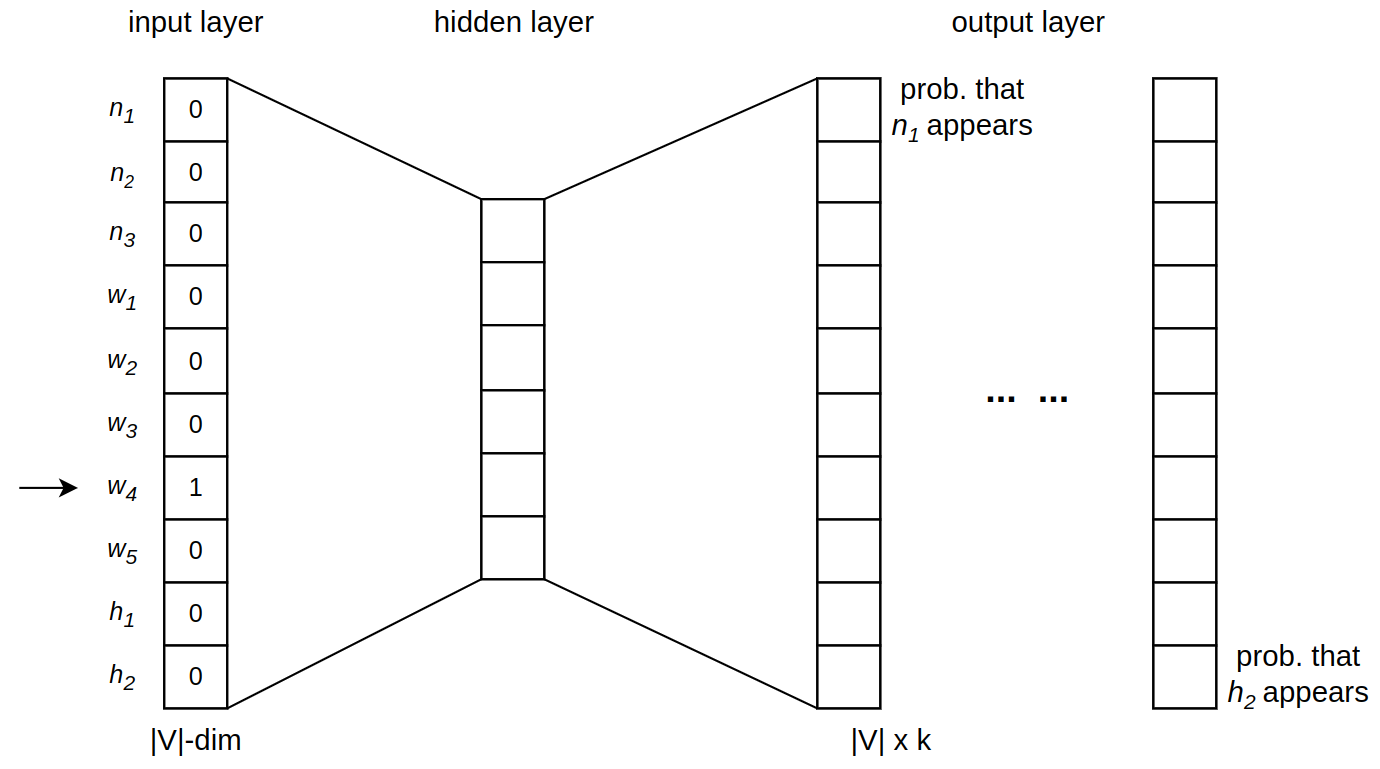}
  \caption{Skip-gram in metapath2vec.}
  \label{fig:skipgram}
\end{subfigure}
\caption{Graph embedding with metapath2vec on WineGraph. Random walks traverse through various paths and gather nodes of different types (sample paths are shown in the left part of the figure).}
\label{fig:metapaths}
\end{figure}

The obtained set of paired food and wine items was utilised to train the metapath2vec model, generating a graph incorporating wines by adding nodes of the wine type to already existing types of nodes in the FlavorGraph.
Incorporating pairings into knowledge graph was performed in the following steps: 1) get top $k$ pairings for each food item ($k = 3$), 2) incorporate wine data into FlavorGraph (nodes), 3) add wine-pairing information to the graph (edges), 4) define new metapaths in FlavorGraph (paths through a heterogeneous graph, (illustrated in Figure~\ref{fig:metapaths}), 5) train FlavorGraph (300-dimensional space). 

\subsection{Experimental Results}
\begin{figure}%
    \centering
    \subfloat{{\includegraphics[width=0.45\textwidth]{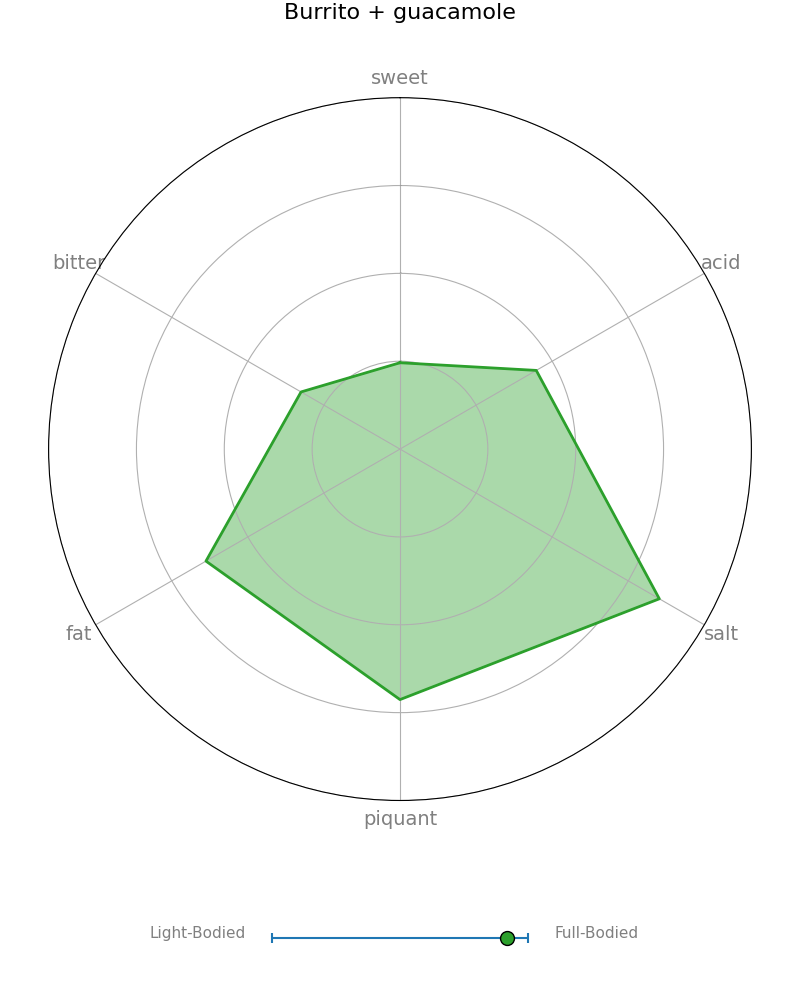} }}%
    \qquad
    \subfloat{{\includegraphics[width=0.45\textwidth]{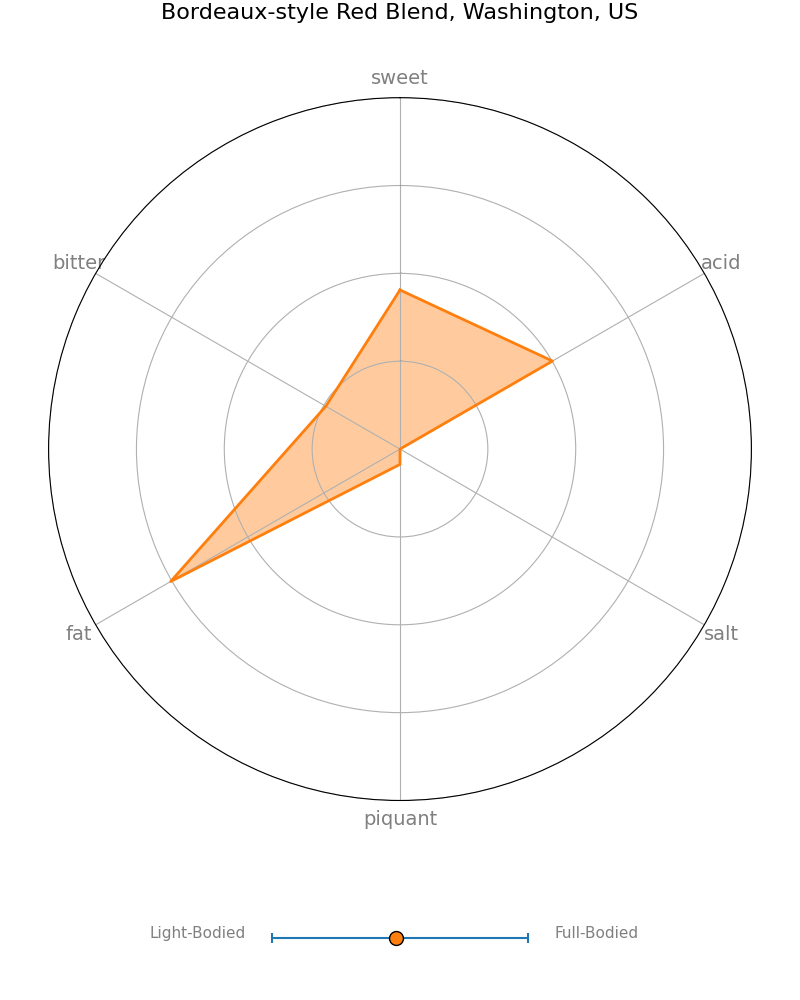} }}%
    \caption{Flavour profiles for wine pairing generated for burrito + guacamole}%
    \label{fig:example}%
%\caption{Sample pairing.}
\end{figure}

\begin{figure}[t]
    \centering
    \includegraphics[scale=0.4]{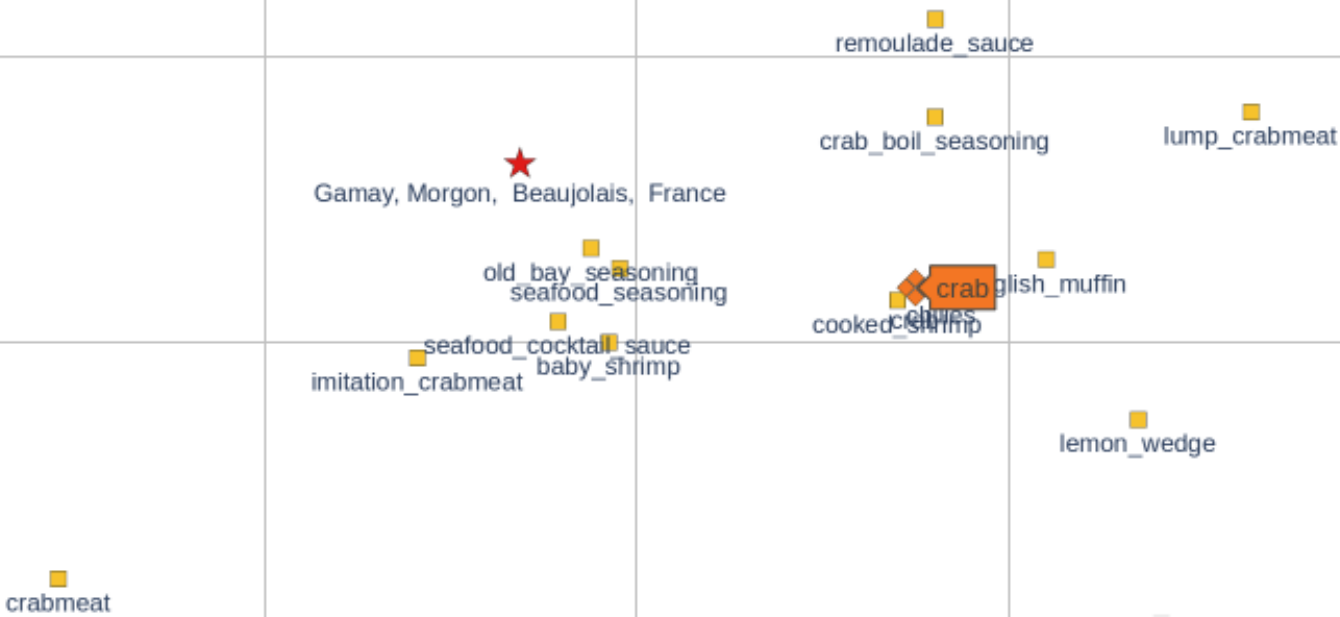}
    \caption{Sample clusters.}
    \label{fig:clusters}
\end{figure}

\vspace{-5pt}
The goal of the experiments was to determine whether wine can be added to the FlavorGraph without loss of quality and while maintaining correct pairings. We have evaluated our method with the use of \emph{Normalised Mutual Information (NMI)} as the quality measure. In our experiments, we wanted to show that for the task of clustering by food category quality is not compromised (see: sample clusters in Figure~\ref{fig:clusters}). In other words, whether NMI is not lower than for the FlavorGraph without wine (see: Table~\ref{tab:training}).
Figure~\ref{fig:example} shows sample flavor profiles. Table~\ref{tab:burrito} shows sample pairings.
\begin{table}[t]
    \caption{Comparison of Normalized Mutual Information (NMI) values for different epochs.}
    \centering
    \begin{tabular}{c|cc}
         \textbf{Dataset} & \textbf{Epochs} &  \textbf{NMI}\\
         \hline
         FlavorGraph & 10 & 0.309\\
         \hline
         \multirow{4}{6em}{FlavorGraph +  wine} & 5 & 0.341\\
         & 10 & 0.319\\
         & 15 & 0.351\\
         & 20 & \textbf{0.358}\\

    \end{tabular}

    \label{tab:training}
\end{table}
Table~\ref{tab:burrito} shows the top 3 pairings for burrito, generated and as the closest nodes in the graph.
\begin{table}[h]
    \centering
     \caption{Pairings for burrito.}
    \begin{tabular}{p{0.07\textwidth}|  p{0.4\textwidth} | p{0.4\textwidth}}
         \textbf{Top} & \textbf{Generated pairing} & \textbf{Pairing from the graph}\\
         \hline
         1 & Bordeaux-style Red Blend, Stellenbosch, South Africa & Malbec-Cabernet Sauvignon, Bordeaux-style Red Blend, Mendoza, Argentina\\
         \hline
         2 & Bordeaux-style Red Blend, Lussac Saint-Émilion, Bordeaux, France & Bordeaux-style Red Blend, Stellenbosch, South Africa \\
         \hline
         3 & Malbec-Cabernet Sauvignon, Bordeaux-style Red Blend, Mendoza, Argentina & Bordeaux-style Red Blend, Listrac-Médoc, Bordeaux, France\\
    \end{tabular}
    \label{tab:burrito}
\end{table}
% For a food item:
% \begin{enumerate}
%     \item Calculate aroma and non-aroma descriptors with the use of the trained word2vec model
%     \item Eliminate wines that do not match the food item (predefined set of rules)
%     \item Find congruent and contrasting wines (predefined set of rules)
%     \item Sort by aroma similarity
% \end{enumerate}
%\section{Related Work} 
%The work closest to ours is by Bender et al.~\cite{DBLP:conf/nips/BenderSKHHHBW23}.
%
%FoodKG\cite{haussmann_foodkg_2019} 
%In related research, Im2recipe utilized word2vec to create food representations from recipe data, while Reciptor proposed a set transformer-based model for obtaining recipe embeddings using a knowledge graph-derived triplet sampling approach. 
%FlavorGraph - general pairing (specific foods) for white wine
\vspace{-10pt}
\section{Conclusions}
In this work, we have shown that Wines can be successfully represented in the form of a graph, enhancing food-wine pairing tasks. For this purpose, we have devised a neural-symbolic method comprised of the embeddings of a heterogeneous graph and rules. 

In the future work, we aim to integrate more characteristics of food and wine, and devise novel embedding methods specifically suited for such data. 
%
% ---- Bibliography ----
%
% BibTeX users should specify bibliography style 'splncs04'.
% References will then be sorted and formatted in the correct style.
%
%\bibliographystyle{splncs04}
%\bibliography{nesy2024}

%
%\begin{thebibliography}{8}
%\bibitem{ref_article1}
%Author, F.: Article title. Journal \textbf{2}(5), 99--110 (2016)
%\end{thebibliography}
\newpage

%\section*{Appendix}

\end{document}